\documentclass[final,3p,times]{elsarticle}

\usepackage{amssymb}
\usepackage{lipsum}
\usepackage{subcaption,tabularx}
\usepackage{makecell}

\journal{Results in Physics}

\begin{document}

\begin{frontmatter}



\title{Synthetic Pattern Generation and Detection of Financial Activities using Graph Autoencoders}


\author[first]{Francesco Zola}
\author[first]{Lucia Muñoz}
\author[first]{Andrea Venturi}
\author[first]{Amaia Gil}
\affiliation[first]{organization={Vicomtech},
            addressline={Mikeletegi Pasealekua, 57}, 
            city={San Sebastian},
            postcode={20009}, 
            country={Spain}}

\begin{abstract}
Illicit financial activities such as money laundering often manifest through recurrent topological patterns in transaction networks. Detecting these patterns automatically remains challenging due to the scarcity of labeled real-world data and strict privacy constraints. To address this, we investigate whether Graph Autoencoders (GAEs) can effectively learn and distinguish topological patterns that mimic money laundering operations when trained on synthetic data. The analysis consists of two phases: (i) data generation, where synthetic samples are created for seven well-known illicit activity patterns using parametrized generators that preserve structural consistency while introducing realistic variability; and (ii) model training and validation, where separate GAEs are trained on each pattern without explicit labels, relying solely on reconstruction error as an indicator of learned structure. We compare three GAE implementations based on three distinct convolutional layers: Graph Convolutional (GAE-GCN), GraphSAGE (GAE-SAGE), and Graph Attention Network (GAE-GAT). Experimental results show that GAE-GCN achieves the most consistent reconstruction performance across patterns, while GAE-SAGE and GAE-GAT exhibit competitive results only in few specific patterns. These findings suggest that graph-based representation learning on synthetic data provides a viable path toward developing AI-driven tools for detecting illicit behaviors, overcoming the limitations of financial datasets.

\end{abstract}




\end{frontmatter}



\section{Introduction}

The detection of money laundering patterns in traditional financial systems is a critical area of research due to the significant economic impact of illicit financial activities \cite{europol2024iocta}. Traditional anti-money laundering (AML) systems rely on predefined rule-based models tailored to flag suspicious activities, such as large cash deposits, frequent international transfers, or transactions involving tax havens \cite{moyes2005raise}. Therefore, criminal \textit{modus operandi} often involve sophisticated laundering schemes designed to evade these rule-based systems, creating complex transaction networks \cite{europol2025socta}. 

In this scenario, identifying, understanding, and modeling transactional patterns created by criminals is essential, as these patterns can guide the development of more effective detection methodologies. This modelization can be done by mapping the transactions into a directed graph where accounts (sender and receiver) are represented as nodes, while the achieved transactions are the edges. Using this graph-based approach, many studies have identified structural topological patterns commonly associated with illicit activities, particularly with money laundering operations \cite{dumitrescu2022anomaly, altman2023realistic}. However, their automatic detection in transactional data remains challenging due to the limited availability of information (caused by privacy and legal constraints) in financial datasets, as well as the lack of structural labels needed to properly train Artificial Intelligence (AI) models. 

For this reason, the aim of this study is to validate whether Graph Machine Learning (GML) models can learn and distinguish patterns that resemble money laundering operations generated synthetically, following specifications reported in the literature \cite{dumitrescu2022anomaly, altman2023realistic,zola2025graphML}. To this end, a two-phase approach is introduced in this work. In the first phase, topological generators are presented. These generators are capable of producing synthetic samples that represent seven well-known patterns commonly associated with illicit activities. These generators allow us to address the concerns related to the data availability and structural labelling. Then, in the second phase, the synthetic samples produced by the generators are used to train and test Graph Autoencoder (GAE) models. Specifically, a separate GAE is trained for each pattern and then tested with samples from all patterns. "This approach relies on the assumption that if a given pattern has been correctly learned during training, the model should exhibit a low reconstruction error for samples of that pattern, while producing higher errors for samples of unknown patterns. Indeed, if this occurs, the GAE and the resulting reconstruction error can be used as a threshold for classifying new samples. In particular, this study analyses the performance of three different GAE implementations, representing a preliminary step toward its usage for detecting topological patterns.

\section{Background}\label{sec:background}
This section introduces background information relevant to the proposed analysis. Specifically, Section~\ref{subsec:pattern} presents topological patterns frequently observed in suspicious activities, while Section~\ref{subsec:related} reviews approaches from related work.

\subsection{Graph Topological Patterns}\label{subsec:pattern}
\begin{figure}[ht]
\centering
  \begin{subfigure}[b]{0.13\linewidth}
   \includegraphics[width=\linewidth]{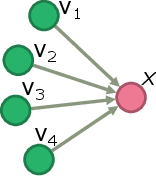}
    \caption{Collector.}
    \label{fig:collector}
  \end{subfigure}
  \hspace{0.3cm}
  \begin{subfigure}[b]{0.16\linewidth}
    \includegraphics[width=\linewidth]{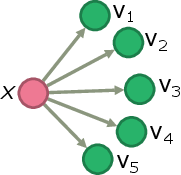}
    \caption{Sink.}
    \label{fig:sink}
  \end{subfigure}
  \hspace{0.3cm}
\begin{subfigure}[b]{0.18\linewidth}
   \includegraphics[width=\linewidth]{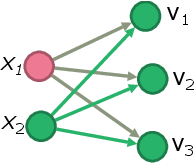}
    \caption{Collusion.}
    \label{fig:collusion}
  \end{subfigure}
  \begin{subfigure}[b]{0.2\linewidth}
   \includegraphics[width=\linewidth]{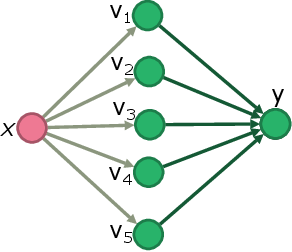}
    \caption{Scatter-Gather.}
    \label{fig:sg}
  \end{subfigure}
  \hspace{0.4cm}
  \begin{subfigure}[b]{0.22\linewidth}
  \centering
    \includegraphics[width=\linewidth]{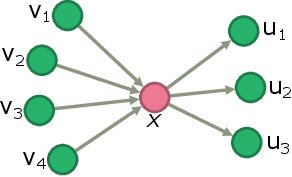}
    \caption{Gather-Scatter.}
    \label{fig:gs}
  \end{subfigure}
      \hspace{0.3cm}
    \begin{subfigure}[b]{0.2\linewidth}
   \includegraphics[width=\linewidth]{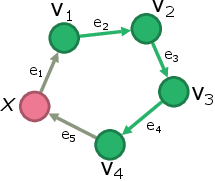}
    \caption{Cyclic.}
    \label{fig:cycle}
  \end{subfigure}  
  \begin{subfigure}[b]{0.2\linewidth}
    \includegraphics[width=\linewidth]{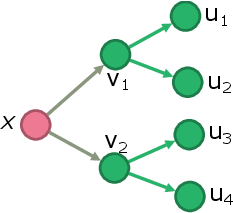}
    \caption{Branching.}
    \label{fig:branch}
  \end{subfigure}
    \hspace{0.4cm}

  \caption{Topological patterns observed in suspicious financial activities.}
  \label{fig:pattern1}
\end{figure}

We now present seven graph topological patterns that frequently appear in suspicious—potentially illicit—financial activities (Table \ref{tab:pattern}), and which are already introduced in different fiat-based and cryptocurrency literature analysis, such as \cite{dumitrescu2022anomaly, altman2023realistic,zola2024unveiling}. In these patterns, extracted from transactional graphs, nodes correspond to financial accounts (senders or receivers), whereas edges are the transactions between them (Figure \ref{fig:pattern1}). 
In particular, these patterns can be divided into three categories: \textit{Single-step, Two-step,} and \textit{Multiple-step}. The first category includes patterns that can be identified by directly examining the incoming or outgoing transactions of a node (\textit{Collector, Sink,} and \textit{Collusion}). That is, starting from a node $X$ and looking at its neighbors in the $1$-step transactional graph. The second category includes two patterns that represent combinations of the three previously introduced patterns and require exactly a $2$-step exploration for their detection (\textit{Scatter-Gather} and \textit{Gather-Scatter}). Finally, the last category includes patterns that unfold over multiple steps, requiring $n$-step graph analysis for their discovery, with $n\geq 2$, such as \textit{Cyclic} and \textit{Branching}.

\begin{table}[]
\begin{tabularx}{\linewidth}{ll X c}
\hline
\multicolumn{1}{c}{\textbf{\#}} & \textbf{Name} & \textbf{Description} & \textbf{Category} \\\hline
\textit{1} &Collector &  This pattern could potentially indicate layering mechanisms in money laundering or the presence of a dormant sender, as drawn in Figure \ref{fig:collector} &Single-step \\
\textit{2} &Sink &  This pattern appears in different fraud, such as smurfing or Ponzi scheme activities (Figure \ref{fig:sink}) &Single-step \\
\textit{3} &Collusion & This pattern suggests a coordinated effort between the input nodes, potentially aimed at concealing the origin of funds, as shown in Figure \ref{fig:collusion} &Single-step \\
\textit{4} & Scatter-Gather (SG) & This pattern represents cases where an input node funds a single recipient through multiple intermediaries, potentially serving as a technique to obscure the origin of funds or evade detection in financial transactions (Figure~\ref{fig:sg}) & Two-step \\
\textit{5} &Gather-Scatter (GS) & This pattern helps identify nodes that operate as a proxy, i.e., nodes that receive and send funds to a similar number of nodes (Figure \ref{fig:gs}) &Two-step\\
\textit{6} &Cyclic &  This pattern highlights (re)cycles in money paths, i.e., if an entity sends money to others and then the money comes back in further steps, as shown in Figure \ref{fig:cycle} &Multiple-step \\
\textit{7} &Branching & This structure forms a recursive splitting pattern, where each level of transaction results in a broader distribution of funds. Such a pattern is frequently indicative of a peeling chain, a common technique used in money laundering (Figure \ref{fig:branch}) &Multiple-step \\
\hline
\end{tabularx}
\caption{Definition of topological patterns related to illicit financial activities.}
\label{tab:pattern}
\end{table}

\subsection{Related Work}\label{subsec:related}
Research on financial crime detection increasingly relies on publicly available datasets, which provide standarized transaction records and synthetic benchmarks for model evaluation \cite{ibm_amlsim, ai4fcf}. Among the most widely used are agent-based simulators such as IBM AMLSim \cite{ibm_amlsim}, PaySim and BankSim \cite{lopez2016paysim}, synthetic markets like Amaretto \cite{labanca2022amaretto} and selected collections such as SAML-D \cite{oztas2023enhancing}, all summarized in recent surveys of open resources \cite{ai4fcf}. These datasets typically deliver tabular transaction logs with sender/receiver identifiers, timestamps, amounts and occasionally typology labels for suspicious activity. While transaction data can naturally be represented as graphs, existing benchmarks are distributed as CSV logs and rarely include explicit topological annotations of illicit subgraphs. For instance, AMLSim outputs histories with per-transaction flags but does not provide motif-level labels \cite{ibm_amlsim}, and SAML-D emphasizes transaction types rather than structural motifs \cite{oztas2023enhancing}. Even more recent efforts in realistic synthetic generation, such as Altman et al. (2023) \cite{altman2023realistic}, focus on reproducing statistical distributions of transactions rather than producing curated graph structures. 
This absence of topology-aware benchmarks limits the assessment of Graph Neural Network (GNN) and embedding methods, which require labeled subgraph patterns to capture relational dynamics in money laundering. Moreover, due to the sensitivity of financial data, real-world labeled datasets remain scarce and rarely shared, reinforcing the importance of synthetic data generation \cite{ai4fcf}. This work directly addresses this gap by generating and labeling topological transaction motifs, enabling GAEs to learn suspicious structures beyond what current datasets allow.

\section{Method for Suspicious Pattern Detection using GML}\label{sec:pattern}

\begin{figure*}[ht]
\centering
  \begin{subfigure}[b]{0.33\linewidth}
   \includegraphics[width=\linewidth]{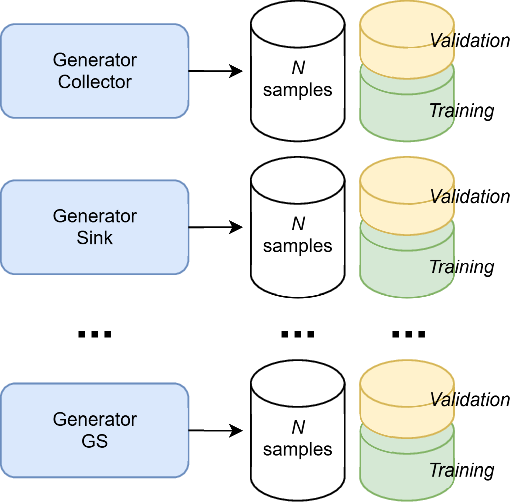}
    \caption{Phase 1: Data generation schema.}
    \label{fig:phase1}
  \end{subfigure}
  \begin{subfigure}[b]{0.65\linewidth}
    \includegraphics[width=\linewidth]{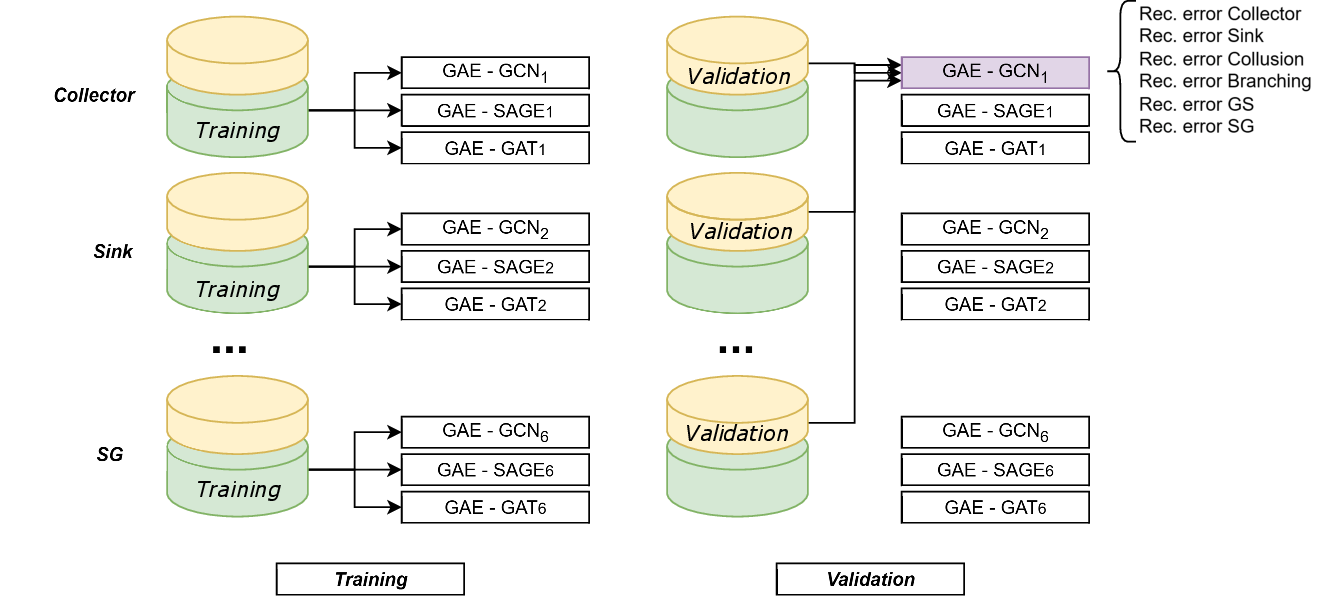}
    \caption{Phase 2: Model training and validation schema.}
    \label{fig:phase2}
  \end{subfigure}
  \caption{Schema of the two phases of the analysis.}
  \label{fig:schema}
\end{figure*}

The analysis consists of two main phases: \textit{Data generation} and \textit{Model training and validation} as illustrated in Figure \ref{fig:schema}. In \textit{Phase 1}, a sufficient number of synthetic samples for each of the seven patterns under consideration is generated (Figure \ref{fig:phase1}). This task involves defining the parameters and ranges used by the synthetic data generators. These generators are designed to replicate and maintain the structural characteristics of their respective patterns while introducing variation in aspects such as the number of nodes and connections, guided by predefined probabilities. For this reason, their parameters are crucial, as they ensure both the structural consistency of each pattern and the necessary variability to reflect a wide range of relational possibilities observed in real-world transactions. Furthermore, since a generator is implemented for each topological pattern, all samples are assigned a corresponding topological label. This approach allows us to address the scarcity of structures and the absence of structural labels. Once the samples for each pattern are generated, they are divided into two sets: training and validation, as shown in Figure \ref{fig:phase1}.
Then, in the \textit{Phase 2}, a GAE model \cite{wu2020comprehensive} is trained for each pattern using the training dataset created in the previous phase (Figure \ref{fig:phase2}). Specifically, each model is trained on data from only one type of pattern. This allows the model to focus on accurately reconstructing the input samples, which inherently belong to the same group (same labels). 

A GAE model consists of two main components: an encoder and a decoder. The encoder is responsible for mapping the input graph data into a lower-dimensional latent space, while the decoder attempts to reconstruct the original input from this latent representation (embeddings). 
In this study, the performance of three GAEs is compared, each one implementing a different convolutional strategy in the encoder component. They are: Graph Convolutional Network (GCN) \cite{kipf2016semi}, GraphSAGE \cite{hamilton2017inductive}, and Graph Attention Network (GAT) \cite{velivckovic2017graph}. From this point forward, we refer to each implementation as GAE-GCN, GAE-SAGE, and GAE-GAT, respectively. 

\section{Data Generation (Phase 1)}\label{sec:data}

Following the schema shown in Figure~\ref{fig:phase1}, $N$ synthetic samples are generated for each pattern. To achieve this, a dedicated graph generator is implemented for each of the considered patterns, following the indication reported in this section. In all cases, the generator uses the pseudo-random Python library \texttt{random} \footnote{https://docs.python.org/3/library/random.html}, ensuring that each execution yields a different graph while respecting specified constraints. This variability contributes to producing a more comprehensive and diverse set of samples.

\textbf{1. Collector:} Starting from a node $X$, the number of incoming nodes ($N_i$) is randomly selected using the function \texttt{random.randint(4,20)}. Then, to introduce additional variability (and noise) into the main structure, $N_o$ nodes are added as outputs to node $X$ if, during execution, the function \texttt{random.random()} returns a value $\leq$ \texttt{0.3}. The number $N_o$ is randomly selected using \texttt{random.randint(1, int($N_i$/3))}.

\textbf{2. Sink:} Starting from a node $X$, the number of outgoing nodes ($N_o$) is randomly selected using the function \texttt{random.randint(4,20)}. Then, to introduce additional variability (and noise) into the main structure, $N_i$ nodes are added as inputs to node $X$ if, during execution, the function \texttt{random.random()} returned a value $\leq$ \texttt{0.3}. The number $N_i$ is randomly selected using \texttt{random.randint(1, int($N_o$/3))}.

\textbf{3. Collusion:} Firstly, the main structure of the collusion is created by defining a number of input nodes ($N_i$) and their respective shared output nodes ($N_o$). Specifically, if the function \texttt{random.random()} returns a value $\leq 0.5$, then 2 input nodes are selected; otherwise, the function \texttt{random.randint(3,4)} is used to determine $N_i$. For the number of shared output nodes ($N_o$), \texttt{random.randint(1,4)} is applied. Each of the $N_i$ nodes is then connected to every $N_o$ node, thereby forming the main structure of the collusion. Furthermore, to introduce additional variability (noise), extra nodes ($N_x$) are added as inputs to the $N_o$ nodes whenever \texttt{random.random()} $\leq 0.3$. The number of such noisy nodes is determined by \texttt{random.randint(1,5)}. The same procedure is repeated to add noisy output nodes to the input $N_i$ nodes.

\textbf{4. SG:} Starting from two nodes $X$ and $Y$, the number of nodes $N_m$, which are outgoing from $X$ and incoming to $Y$, is randomly selected using the function \texttt{random.randint(4,10)}. This creates the main structure of the SG pattern. Then, to introduce additional variability (and noise) into the pattern, a node is added as an input to the $N_m$ nodes only if, during execution, the function \texttt{random.random()} returned a value $\leq$ \texttt{0.2}. This procedure is repeated for the output noisy nodes as well.

\textbf{5. GS:} Starting from the node $X$, the number of incoming nodes $N_i$ is randomly selected using the function \texttt{random.randint(4,10)}. Then, the number of outgoing nodes of $X$ ($N_o$) is established by the function \texttt{random.randint ($N_i$-3, $N_i$+3)} if $N_i \geq 8$, otherwise by the function \texttt{random.randint(3, 7)}. This creates the main structure of the GS pattern. Then, to introduce additional variability (and noise) into the pattern, a node is added as an \textit{output} to the $N_i$ incoming nodes only if, during execution, the function \texttt{random.random()} returned a value $\leq$ \texttt{0.2}. At the same time, a node is added as an \textit{intput} to the $N_o$ incoming nodes only if, during execution, the function \texttt{random.random()} returned a value $\leq$ \texttt{0.2}.

\textbf{6. Cyclic:} Starting from a node $X$, the number of nodes to create the cycle is randomly selected using the function \texttt{random.randint(2,10)}. Then, for each node within the cycle, noisy nodes are added as follows: a number of noisy nodes is selected using \texttt{random.randint(1,2)}, and added to the structure only if, during execution, the function \texttt{random.random()} returns a value $\leq$ \texttt{0.3}. This procedure is repeated for both input and output noisy nodes.

\textbf{7. Branching:} Starting from a node $X$, the number of outgoing nodes is randomly selected using the function \texttt{random.randint(2,3)}. Then, for each of these nodes, the number of their outgoing nodes is determined by the following rules: if \texttt{random.random()} returned a value $\leq$ \texttt{0.08}, 3 outgoing nodes are added; if it returns a value $\leq$ \texttt{0.15}, 0 outgoing nodes are added; otherwise, 2 outgoing nodes are added. This creates the main structure of the Branching pattern.

For the analysis and the validation proposed in this work, 15,000 synthetic samples are generated for each pattern. Each set is then split into 80\% for training and 20\% for validation. The first set (12,000 samples per pattern) is used to train the GAE models, while the remaining 3,000 samples from each pattern are used to validate their performance.

\section{Model Training and Validation (Phase 2)}\label{sec:data}


In this analysis, the three GAE variants (GAE-GCN, GAE-SAGE,
and GAE-GAT) are implemented following the specification reported in \cite{zola2025graphML}. Furthermore, following the indications of the authors, the following nine node properties are extracted and used: \textit{in-degree, out-degree, closeness, betweenness, harmonic centrality, second order centrality, laplacian centrality, constraint,} and \textit{reciprocity}.
These node properties, together with the adjacency matrix, are used as input for the GAE models. Yet, each model is trained for a maximum of 100 epochs, with early stopping set to 3 steps, and a batch size of 25. The Adam optimisation function is used.

\begin{figure}[]
\centering
  \begin{subfigure}[b]{0.32\linewidth}
   \includegraphics[width=\linewidth]{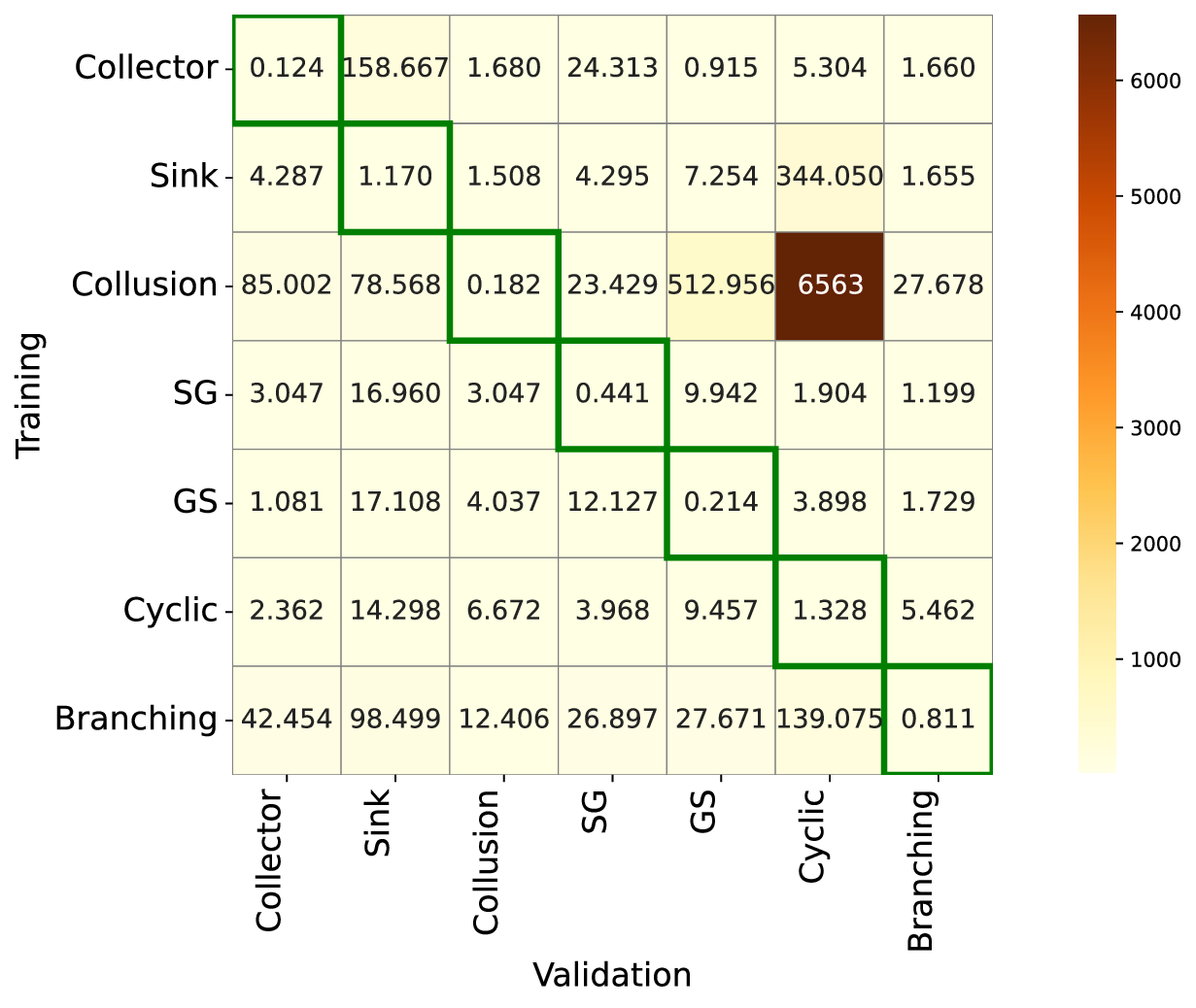}
    \caption{GAE-GCN.}
    \label{fig:gcn}
  \end{subfigure}
  \begin{subfigure}[b]{0.32\linewidth}
    \includegraphics[width=\linewidth]{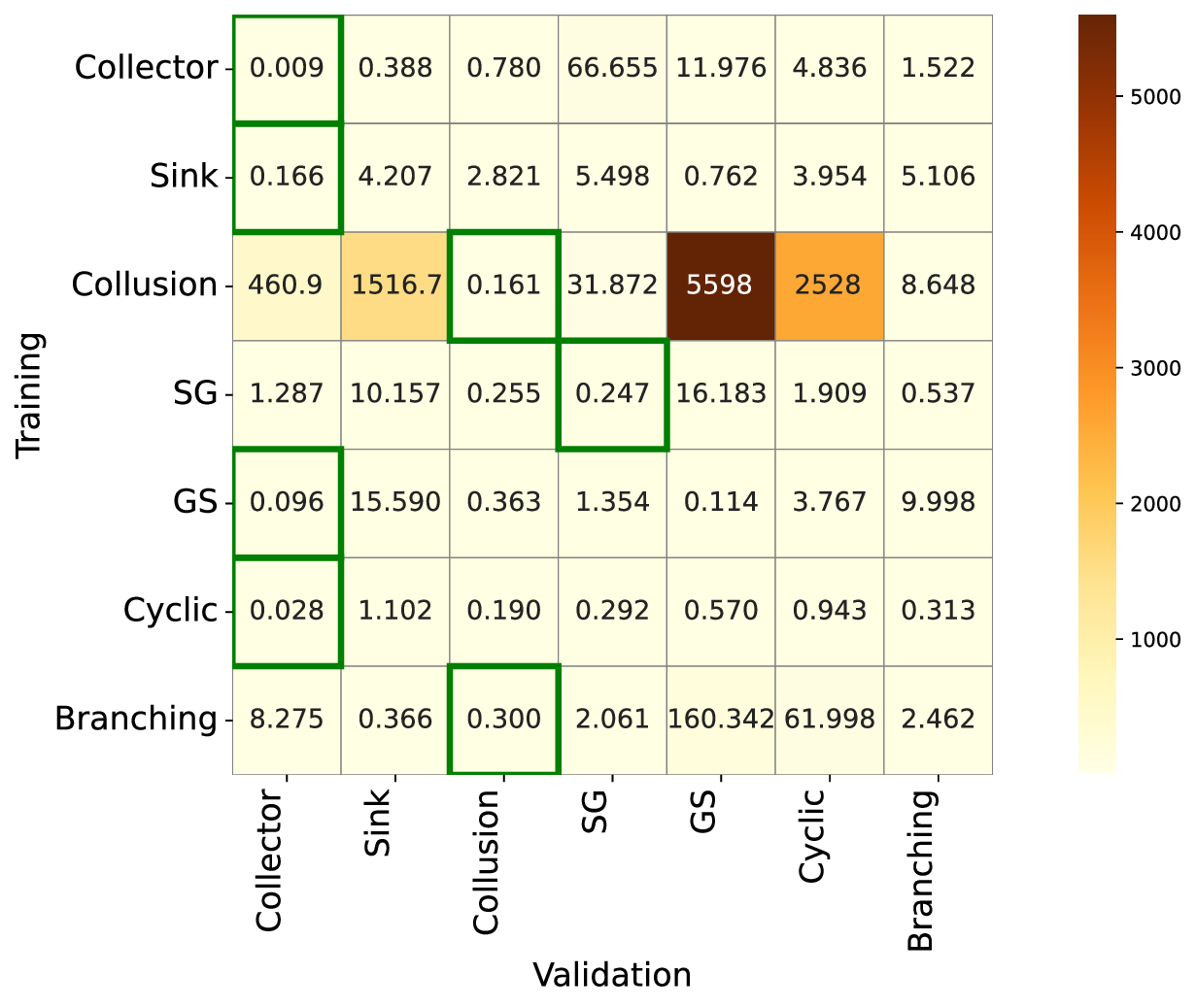}
    \caption{GAE-SAGE.}
    \label{fig:sage}
  \end{subfigure}
    \begin{subfigure}[b]{0.32\linewidth}
    \includegraphics[width=\linewidth]{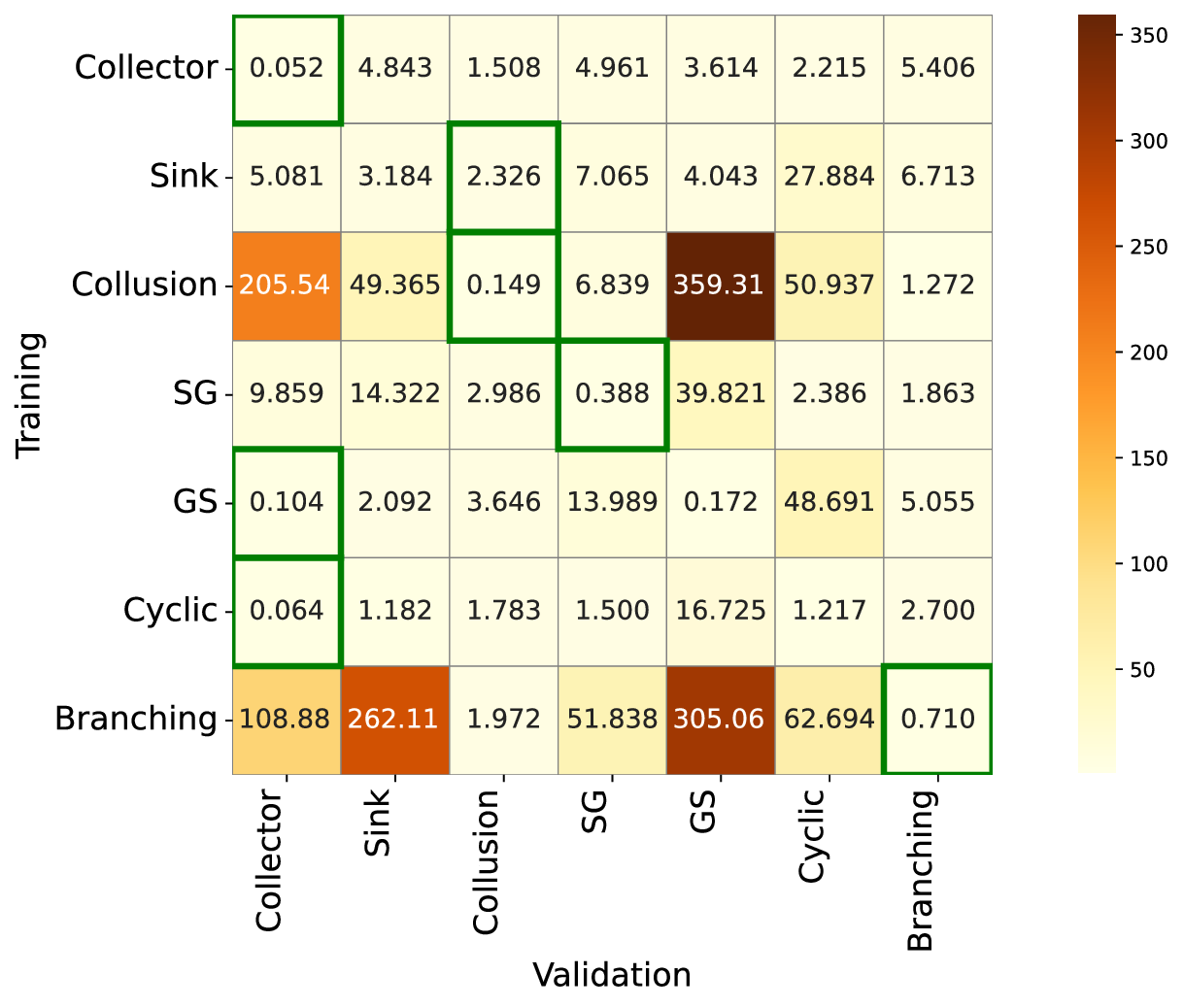}
    \caption{GAE-GAT.}
    \label{fig:gat}
  \end{subfigure}
  \caption{Reconstruction error matrix for the three GAE variants in the validation stage. The lowest error for each trained model is highlighted with a green box.}
  \label{fig:loss}
\end{figure}

Figure \ref{fig:loss} reports the reconstruction error of each GAE model for each pattern in the validation stage. Specifically, the figure shows that GAE-GCN models are the only ones able to reconstruct the trained pattern with the lowest reconstruction error, as indicated by the green boxes on the diagonal of Figure \ref{fig:gcn}. In contrast, GAE-SAGE and GAE-GAT correctly detect patterns in 3 and 4 cases, respectively. Furthermore, considering the overall best model for each structure—that is, the lowest reconstruction error across all variants—Figure \ref{fig:loss} shows that GAE-SAGE performs best for the \textit{Collector} and \textit{SG} patterns, GAE-GAT for the \textit{Collusion} and \textit{Branching} patterns, and GAE-GCN for all others.

\section{Conclusion and Future Work}\label{sec:conclusions}
This work represents a preliminary step toward the use of GAE models for detecting topological patterns potentially related to illicit financial activities. The study aims to address common problems associated with financial datasets—such as legal and privacy concerns and the lack of structurally labeled information—through the use of synthetic topological generators. The analysis shows that GAEs trained with synthetic structure are able to reconstruct, with low error, the elements of the patterns they were trained on, thereby enabling their use as classifiers (by defining a discriminative threshold). As future work, these models could be tested with data from financial datasets such as SAML-D, AMLSim, or Amaretto, which, although synthetically generated, can help validate the generalization of the trained models. By characterizing the distinctive traits of the topological patterns potentially related to illicit activities, it becomes possible to design more resilient monitoring and analysis tools.

\section*{Acknowledgment}
This work has been partially supported by the European Union's Horizon 2020 Research and Innovation Program under the project CEDAR (Grant Agreement No. 101135577) and FALCON (Grant Agreement No. 101121281). The content of this article does not reflect the official opinion of the European Union. Responsibility for the information and views expressed therein lies entirely with the authors.

\bibliographystyle{elsarticle-num}
\bibliography{example}






\end{document}